\documentclass[11pt]{article}
\pdfoutput=1

\usepackage{hyperref}
\usepackage[dvipsnames]{xcolor}
\usepackage[colorinlistoftodos,textsize=xsmall]{todonotes}
\usepackage[utf8]{inputenc}
\usepackage{acl2017}
\usepackage{latexsym}
\usepackage{times}
\usepackage{url}
\usepackage{latexsym}
\usepackage[english]{babel}
\usepackage{amssymb}
\usepackage{amsmath}
\usepackage{graphicx}
\usepackage{wrapfig}
\usepackage{bm}
\usepackage{enumerate}
\usepackage{booktabs}
\usepackage{comment}
\usepackage{multirow}
\usepackage{cleveref}
\crefformat{section}{\S#2#1#3}
\usepackage{subcaption}
\usepackage{mathtools, cuted}

\newcommand\given[1][]{\:#1\vert\:}

\newcommand{\red}[1]{\textcolor{Red}{#1}}
\newcommand{\redbf}[1]{\textbf{\textcolor{Red}{#1}}}

\newcommand{\green}[1]{\textcolor{OliveGreen}{#1}}
\newcommand{\greenbf}[1]{\textbf{\textcolor{OliveGreen}{#1}}}

\newcommand{\whitebf}[1]{\textbf{\textcolor{White}{#1}}}

\aclfinalcopy 
\def\aclpaperid{***} 

\title{Doubly-Attentive Decoder for Multi-modal Neural Machine Translation}

\author{Iacer Calixto \\
  ADAPT Centre\\
  School of Computing\\
  Dublin City University\\
  Dublin, Ireland\\
  \\
  {\tt {\small iacer.calixto@adaptcentre.ie}}\\
  \And
  Qun Liu \\
  ADAPT Centre\\
  School of Computing\\
  Dublin City University\\
  Dublin, Ireland\\
  \\
  \And
  Nick Campbell \\
  ADAPT Centre\\
  Speech Communication Lab\\
  Trinity College Dublin\\
  Dublin 2, Ireland\\
  \\
  }

\date{}

\begin{document}
\maketitle
\begin{abstract}
 We introduce a Multi-modal Neural Machine Translation model in which a doubly-attentive decoder naturally incorporates \emph{spatial} visual features obtained using pre-trained convolutional neural networks, bridging the gap between image description and translation.
 Our decoder learns to attend to source-language words and parts of an image independently by means of two \emph{separate attention mechanisms} as it generates words in the target language.
 We find that our model can efficiently exploit not just back-translated in-domain multi-modal data but also large general-domain text-only MT corpora.
 We also report state-of-the-art results on the Multi30k data set.
\end{abstract}

\section{Introduction}\label{sec:intro}

Neural Machine Translation (NMT) has been successfully tackled as a \emph{sequence to sequence} learning problem~\cite{KalchbrennerBlunsom2013,Choetal2014,SutskeverVinyalsLe2014} where each training example consists of one source and one target variable-length sequences, with no prior information on the alignment between the two.

In the context of NMT, \newcite{BahdanauChoBengio2015} first proposed to use an \emph{attention mechanism} in the decoder, which is trained to attend to the relevant source-language words as it generates each word of the target sentence.
Similarly, ~\newcite{Xuetal2015} proposed an attention-based model for the task of image description generation (IDG) where a model learns to attend to specific parts of an image representation (the source) as it generates its description (the target) in natural language.

We are inspired by recent successes in applying attention-based models to NMT and IDG.
In this work, we propose an end-to-end attention-based multi-modal neural machine translation (MNMT) model which effectively incorporates \emph{two independent attention mechanisms}, one over source-language words and the other over different areas of an image.

Our main contributions are:
\begin{itemize}
 \setlength\itemsep{0em}
 \item We propose a novel attention-based MNMT model which incorporates spatial visual features in a separate visual attention mechanism;
 
 \item We use a medium-sized, back-translated multi-modal in-domain data set and large general-domain text-only MT corpora to pre-train our models and show that our MNMT model can efficiently exploit them;
 
 \item We show that images bring useful information into an NMT model, in situations in which sentences describe objects illustrated in the image.
\end{itemize}

To the best of our knowledge, previous MNMT models in the literature that utilised spatial visual features did not significantly improve over a comparable model that used global visual features or even only textual features~\cite{Caglayanetal2016,CalixtoElliottFrank2016,Huangetal2016,Libovickyetal2016,Speciaetal2016}.
In this work, we wish to address this issue and propose an MNMT model that uses, in addition to an attention mechanism over the source-language words, an additional visual attention mechanism to incorporate spatial visual features, and still improves on simpler text-only and multi-modal attention-based NMT models.

The remainder of this paper is structured as follows.
We first briefly revisit the attention-based NMT framework (\cref{sec:background}) and expand it into an MNMT framework (\cref{sec:multimodal-model}).
In \cref{sec:dataset}, we introduce the datasets we use to train and evaluate our models,
in~\cref{sec:experiments} we discuss our experimental setup and analyse and discuss our results.
Finally, in~\cref{sec:related} we discuss relevant previous related work and in~\cref{sec:conclusions} we draw conclusions and provide some avenues for future work.

\section{Background and Notation}
\label{sec:background}

\subsection{Attention-based NMT}\label{sec:model}

We describe the attention-based NMT model introduced by \newcite{BahdanauChoBengio2015} in this section.
Given a source sequence ${X = (x_1, x_2, \cdots, x_N)}$ and
its translation ${Y = (y_1, y_2, \cdots, y_M)}$,
an NMT model aims to build a single neural network that translates $X$ into $Y$ by directly learning to model $p(Y \given X)$.
The entire network consists of one \emph{encoder} and one \emph{decoder} with one \emph{attention mechanism}, typically implemented as two Recurrent Neural Networks (RNN) and one multilayer perceptron, respectively.
Each $x_i$ is a row index in a source lookup or word embedding matrix
$\bm{E_x} \in \mathbb{R}^{|V_x| \times d_x}$,
as well as each $y_j$ being an index in a target lookup or word embedding matrix
$\bm{E_y} \in \mathbb{R}^{|V_y| \times d_y}$,
$V_x$ and $V_y$ are source and target vocabularies, and $d_x$ and $d_y$ are source and target word embeddings dimensionalities, respectively.

The encoder is a bi-directional RNN with GRU~\cite{Choetal2014b}, where a forward RNN $\overrightarrow{\Phi}_{\text{enc}}$ reads $X$ word by word, from left to right, and generates a sequence of \emph{forward annotation vectors} ${(\overrightarrow{\bm{h}}_1, \overrightarrow{\bm{h}}_2, \cdots, \overrightarrow{\bm{h}}_N)}$ at each encoder time step ${i \in [1,N]}$.
Similarly, a backward RNN $\overleftarrow{\Phi}_{\text{enc}}$ reads $X$ from right to left, word by word, and generates a sequence of \emph{backward annotation vectors} ${(\overleftarrow{\bm{h}}_N, \overleftarrow{\bm{h}}_{N-1},\cdots, \overleftarrow{\bm{h}}_1)}$.
The final annotation vector is the concatenation of forward and backward vectors $\bm{h}_i = \big[ \overrightarrow{\bm{h}_i}; \overleftarrow{\bm{h}_i} \big]$, and ${C = (\bm{h}_1, \bm{h}_2, \cdots, \bm{h}_N)}$ is the set of source annotation vectors.

These annotation vectors are in turn used by the decoder, which is essentially a neural language model (LM)~\cite{Bengioetal2003} conditioned on the previously emitted words and the source sentence via an attention mechanism.
A multilayer perceptron is used to initialise the decoder's hidden state $\bm{s}_0$ at time step $t=0$, where the input to this network is the concatenation of the last forward and backward vectors $\big[ \overrightarrow{\bm{h}_N}; \overleftarrow{\bm{h}_1} \big]$.

At each time step $t$ of the decoder, a \emph{time-dependent} source context vector $\bm{c}_t$ is computed based on the annotation vectors $C$ and the decoder previous hidden state $\bm{s}_{t-1}$.
This is part of the formulation of the \textit{conditional GRU} and is described further in~\cref{sec:sentence_context}.
In other words, the encoder is a bi-directional RNN with GRU and the decoder is an RNN with a conditional GRU.

Given a hidden state $\bm{s}_t$,
the probabilities for the next target word
are computed using one projection layer followed by a softmax layer as described in \cref{eq:logit},
where the matrices $\bm{L}_o$, $\bm{L}_s$, $\bm{L}_w$ and $\bm{L}_c$ are transformation matrices and $\bm{c}_t$ is a time-dependent source context vector generated by the conditional GRU.

\begin{figure*}[ht!]\vspace{-10px}
\begin{equation}
  \label{eq:logit}
  p(y_t = k \given \bm{y_{<t}}, \bm{c_t}) =
  \text{Softmax}(\bm{L_o} \tanh( \bm{L_s} \bm{s}_t +
                        \bm{L_w} \bm{E_y}[\hat{y}_{t-1}] +
                        \bm{L_c} \bm{c_t} ) ).
\end{equation}
\end{figure*}

\subsection{Conditional GRU}
\label{sec:sentence_context}

The conditional GRU\footnote{\url{https://github.com/nyu-dl/dl4mt-tutorial/blob/master/docs/cgru.pdf}.} has three main components computed at each time step $t$ of the decoder:
\begin{itemize}
  \setlength\itemsep{0em}
  \item REC$_1$ computes a hidden state proposal $\bm{s}'_t$ based on the previous hidden state $\bm{s}_{t-1}$ and the previously emitted word $\hat{y}_{t-1}$;
  
  \item ATT$_\text{src}$\footnote{ATT$_\text{src}$ is named ATT in the original technical report.} is an attention mechanism over the hidden states of the source-language RNN and computes $\bm{c}_t$ using all source annotation vectors $\bm{C}$ and the hidden state proposal $\bm{s}'_t$;
  
  \item REC$_2$ computes the final hidden state $\bm{s}_t$ using the hidden state proposal $\bm{s}'_t$ and the time-dependent source context vector $\bm{c}_t$.
\end{itemize}

We use the conditional GRU in our text-only attention-based NMT model.
First, a single-layer feed-forward network is used to compute an \emph{expected alignment} $\bm{e}^\text{src}_{t,i}$ between each source annotation vector $\bm{h}_i$ and the target word $\hat{y}_t$ to be emitted at the current time step $t$, as shown in Equations~(\ref{eq:expected-alignment-src}) and~(\ref{eq:alpha-src}):
\begin{align}
\bm{e}^\text{src}_{t,i} &= (\bm{v}^\text{src}_a)^T \tanh( \bm{U}^\text{src}_a \bm{s}'_t + \bm{W}^\text{src}_a \bm{h}_i),\label{eq:expected-alignment-src}\\
\bm{\alpha}^\text{src}_{t,i} &= \frac{\exp{(\bm{e}^\text{src}_{t,i})}}{ \sum_{j=1}^{N}{\exp{(\bm{e}^\text{src}_{t,j})}} },\label{eq:alpha-src}
\end{align}
\noindent
where $\bm{\alpha}^\text{src}_{t,i}$ is the normalised alignment matrix between each source annotation vector $\bm{h}_i$ and the word $\hat{y}_t$ to be emitted at time step $t$, and $\bm{v}^\text{src}_a$, $\bm{U}^\text{src}_a$ and $\bm{W}^\text{src}_a$ are model parameters.

Finally, a time-dependent source context vector $\bm{c}_t$ is computed as a weighted sum over the source annotation vectors, where each vector is weighted by the attention weight $\bm{\alpha}^\text{src}_{t,i}$, as in~\cref{eq:time-dependent-annotation-vector}:
\begin{equation}
\label{eq:time-dependent-annotation-vector}
    \bm{c}_t = \text{$\sum_{i=1}^{N}{ \bm{\alpha}^\text{src}_{t,i} \bm{h}_i }$}.
\end{equation}

\section{Multi-modal NMT}
\label{sec:multimodal-model}

\begin{figure}[t!]
 \centering
 \includegraphics[width=0.48\textwidth]{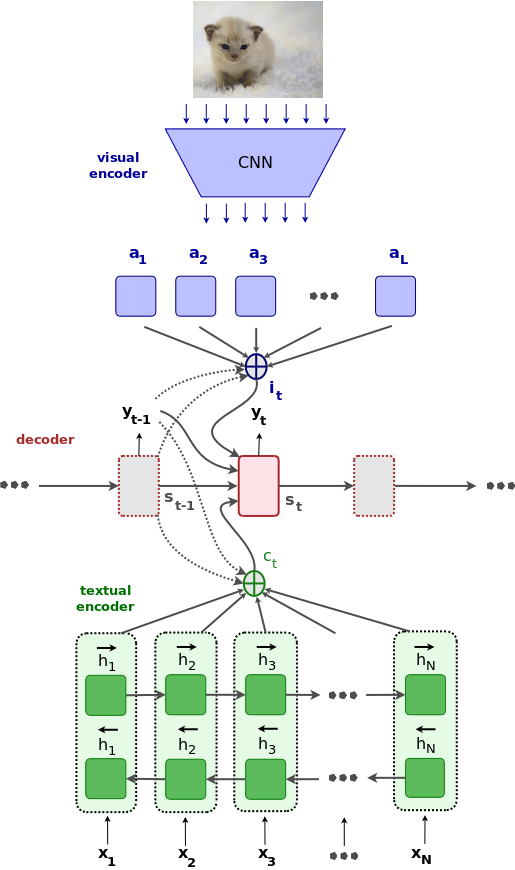}
 \caption{
 A doubly-attentive decoder learns to attend to image patches and source-language words independently when generating translations.
 }
 \label{fig:separate-doubly-attentive}
\end{figure}

Our MNMT model can be seen as an expansion of the attention-based NMT framework described in \cref{sec:model} with the addition of a \emph{visual component} to incorporate spatial visual features, and is comparable to the model evaluated by \newcite{CalixtoElliottFrank2016}.

We use publicly available pre-trained CNNs for image feature extraction.
Specifically, we extract spatial image features for all images in our dataset using the 50-layer Residual network (ResNet-50) of \newcite{He2015}.
These spatial features are the activations of the \texttt{res4f} layer, which can be seen as encoding an image in a 14$\times$14 grid, where each of the entries in the grid is represented by a 1024D feature vector that only encodes information about that specific region of the image.
We vectorise this 3-tensor into a 196$\times$1024 matrix $A = (\bm{a}_1, \bm{a}_2, \cdots, \bm{a}_L), \bm{a}_l \in \mathbb{R}^{1024}$ where each of the L $=196$ rows consists of a 1024D feature vector and each column, i.e. feature vector, represents one grid in the image.

\subsection{\texorpdfstring{NMT$_{\text{SRC+IMG}}$}{}: decoder with two independent attention mechanisms}
\label{sec:model_2}

\begin{figure*}[ht!]\vspace{-10px}
\begin{equation}\label{eq:logit_imgrnn}
  p(y_t = k \given \bm{y_{<t}}, C, A) =
    softmax(\bm{L_o} \tanh(
    \bm{L_{s}} \bm{s}_t +
    \bm{L_w} \bm{E_y}[\hat{y}_{t-1}] +
    \bm{L_{cs}} \bm{c}_t +
    \bm{L_{ci}} \bm{i}_t)),
\end{equation}
\end{figure*}

Model NMT$_{\text{SRC+IMG}}$ integrates two separate attention mechanisms over the source-language words and visual features in a single decoder RNN.
Our doubly-attentive decoder RNN is conditioned on the previous hidden state of the decoder and the previously emitted word, as well as the source sentence and the image via two independent attention mechanisms, as illustrated in Figure~\ref{fig:separate-doubly-attentive}.

We implement this idea expanding the conditional GRU described in \cref{sec:sentence_context} onto a \emph{doubly-conditional} GRU.
To that end, in addition to the source-language attention, we introduce a new attention mechanism ATT$_\text{img}$ to the original conditional GRU proposal.
This visual attention computes a \emph{time-dependent} image context vector $\bm{i}_t$ given a hidden state proposal $\bm{s}'_t$ and the image annotation vectors $A = (\bm{a}_1, \bm{a}_2, \cdots, \bm{a}_L)$ using the ``soft'' attention \cite{Xuetal2015}.

This attention mechanism is very similar to the source-language attention with the addition of a \textit{gating scalar}, explained further below.
First, a single-layer feed-forward network is used to compute an \emph{expected alignment} $\bm{e}^\text{img}_{t,l}$ between each image annotation vector $\bm{a}_l$ and the target word to be emitted at the current time step $t$, as in~\cref{eq:expected-alignment-img,eq:alpha-img}:
\begin{align}
\bm{e}^\text{img}_{t,l} &= (\bm{v}^\text{img}_a)^T \tanh( \bm{U}^\text{img}_a \bm{s}'_t + \bm{W}^\text{img}_a \bm{a}_l),\label{eq:expected-alignment-img}\\
\bm{\alpha}^\text{img}_{t,l} &= \frac{\exp{(\bm{e}^\text{img}_{t,l})}}{ \sum_{j=1}^{L}{\exp{(\bm{e}^\text{img}_{t,j})}}\label{eq:alpha-img} },
\end{align}
\noindent
where $\bm{\alpha}^\text{img}_{t,l}$ is the normalised alignment matrix between all the image patches $\bm{a}_l$ and the target word to be emitted at time step $t$, and $\bm{v}^\text{img}_a$, $\bm{U}^\text{img}_a$ and $\bm{W}^\text{img}_a$ are model parameters.
Note that Equations~(\ref{eq:expected-alignment-src}) and (\ref{eq:alpha-src}), that compute the expected source alignment $\bm{e}^\text{src}_{t,i}$ and the weight matrices $\bm{\alpha}^\text{src}_{t,i}$, and \cref{eq:expected-alignment-img,eq:alpha-img} that compute the expected image alignment $\bm{e}^\text{img}_{t,l}$ and the weight matrices $\bm{\alpha}^\text{img}_{t,l}$, both compute similar statistics over the source and image annotations, respectively.

In \cref{eq:beta} we compute $\beta_t \in [0,1]$, a gating scalar used to weight the expected importance of the image context vector in relation to the next target word at time step $t$:
\begin{equation}\label{eq:beta}
  \beta_t = \sigma( \bm{W}_{\beta} \bm{s}_{t-1} + \bm{b}_{\beta}),
\end{equation}
\noindent
where $\bm{W}_{\beta}$, $\bm{b}_{\beta}$ are model parameters.
It is in turn used to compute the time-dependent image context vector $\bm{i}_t$ for the current decoder time step $t$, as in~\cref{eq:time-dependent-image-annotation-vector}:
\begin{equation}\label{eq:time-dependent-image-annotation-vector}
  \bm{i}_t = \beta_t \sum_{l=1}^{L}{ \bm{\alpha}^\text{img}_{t,l} \bm{a}_l }.
\end{equation}

The only difference between Equations~(\ref{eq:time-dependent-annotation-vector}) (source context vector) and~(\ref{eq:time-dependent-image-annotation-vector}) (image context vector) is that the latter uses a gating scalar, whereas the former does not.
We use $\beta$ following~\newcite{Xuetal2015} who empirically found it to improve the variability of the image descriptions generated with their model.

Finally, we use the time-dependent image context vector $\bm{i}_t$ as an additional input to a modified version of REC$_2$ (\cref{sec:sentence_context}), which now computes the final hidden state $\bm{s}_t$ using the hidden state proposal $\bm{s}'_t$, and the time-dependent source and image context vectors $\bm{c}_t$ and $\bm{i}_t$, as in (\ref{eq:gru_imga}):
\begin{align}
\label{eq:gru_imga}
\bm{z}_t &= \sigma (\bm{W}^\text{src}_z \bm{c}_t + \bm{W}^\text{img}_z \bm{i}_t + \bm{U}_z \bm{s}'_j),\notag\\
\bm{r}_t &= \sigma (\bm{W}^\text{src}_r \bm{c}_t + \bm{W}^\text{img}_r \bm{i}_t + \bm{U}_r \bm{s}'_j),\notag\\
\underline{\bm{s}}_t &= \tanh( \bm{W}^\text{src} \bm{c}_t + \bm{W}^\text{img} \bm{i}_t + \bm{r}_t \odot (\bm{U}\bm{s}'_t)),\notag\\
\bm{s}_t &= (1 - \bm{z}_t)\odot \underline{\bm{s}}_t + \bm{z}_t \odot \bm{s}'_t.
\end{align}

In Equation (\ref{eq:logit_imgrnn}), the probabilities for the next target word are computed using the new multi-modal hidden state $\bm{s}_t$, the previously emitted word $\hat{y}_{t-1}$, and the two context vectors $\bm{c}_t$ and $\bm{i}_t$, where $\bm{L}_o$, $\bm{L}_s$, $\bm{L}_w$, $\bm{L}_{cs}$ and $\bm{L}_{ci}$ are projection matrices and trained with the model.

\section{Data}
\label{sec:dataset}

The Flickr30k data set contains 30k images and 5 descriptions in English for each image~\cite{Youngetal2014}.
In this work, we use the Multi30k dataset \cite{ElliottFrankSimaanSpecia2016}, which consists of two multilingual expansions of the original Flickr30k: one with translated data and another one with comparable data, henceforth referred to as M30k$_\text{T}$ and M30k$_\text{C}$, respectively.

For each of the 30k images in the Flickr30k, the M30k$_\text{T}$ has one of the English descriptions manually translated into German by a professional translator. Training, validation and test sets contain 29k, 1,014 and 1k images respectively, each accompanied by a sentence pair (the original English sentence and its translation into German).
For each of the 30k images in the Flickr30k, the M30k$_\text{C}$ has five descriptions in German collected independently from the English descriptions. Training, validation and test sets contain 29k, 1,014 and 1k images respectively, each accompanied by five sentences in English and five sentences in German.

We use the entire M30k$_\text{T}$ training set for training our MNMT models, its validation set for model selection with BLEU~\cite{Papinenietal2002}, and its test set for evaluation.
In addition, since the amount of training data available is small, we build a back-translation model using the text-only NMT model described in~\cref{sec:model} trained on the Multi30k$_\text{T}$ data set (German$\rightarrow$English), without images.
We use this model to back-translate the $145$k German descriptions in the Multi30k$_\text{C}$ into English and include the triples (synthetic English description, German description, image) as additional training data~\cite{Sennrichetal2016a}.

We also use the WMT 2015 text-only parallel corpora available for the English--German language pair, consisting of about $4.3$M sentence pairs ~\cite{Bojaretal2015}.
These include the Europarl v7~\cite{Koehn2005}, News Commentary and Common Crawl corpora, which are concatenated and used for pre-training.

We use the scripts in the Moses SMT Toolkit~\cite{Koehnetal2007} to normalise and tokenize English and German descriptions, and we also convert space-separated tokens into subwords~\cite{Sennrichetal2016}.
All models use a common vocabulary of $83,093$ English and $91,141$ German subword tokens.
If sentences in English or German are longer than 80 tokens, they are discarded.
We train models to translate from English into German and report evaluation of cased, tokenized sentences with punctuation.

\section{Experimental setup}
\label{sec:experiments}

\begin{table*}[t!]
  \centering
  \resizebox{0.8\linewidth}{!} {
  \begin{tabular}{lllllll}
    
    \toprule
    \textbf{Model} &
    \textbf{Training} &
    \textbf{BLEU4}$\uparrow$ &
    \textbf{METEOR}$\uparrow$ &
    \textbf{TER}$\downarrow$ &
    \multicolumn{2}{l}{\textbf{chrF3}$\uparrow$ (prec. / recall)} \\
    
    &
    \textbf{data} &
    &
    &
    &
    \\
    \midrule
    
    NMT &
    M30k$_\text{T}$ &
    \underline{33.7} &
    52.3 &
    46.7 &
    65.2 & (67.7 / 65.0)\\
     
    PBSMT &
    M30k$_\text{T}$ &
    32.9 &
    \underline{54.3}$^\dagger$ &
    \underline{45.1}$^\dagger$ &
    \textbf{\underline{67.4}} & (66.5 / 67.5) \\
    
    \newcite{Huangetal2016} &
    M30k$_\text{T}$ &
    35.1 {\small \greenbf{($\uparrow1.4$)}} &
    52.2 {\small \redbf{($\downarrow2.1$)}} &
    --- &
    --- &
    --- \\
     
    & + RCNN &
    \textbf{36.5} {\small \greenbf{($\uparrow2.8$)}} &
    54.1 {\small \redbf{($\downarrow0.2$)}} &
    --- &
    --- &
    --- \\
    \midrule
    
    NMT$_{\text{SRC+IMG}}$ &
    M30k$_\text{T}$ &
    \textbf{36.5}$^{\dagger\ddagger}$ &
    \textbf{55.0}$^\dagger$  &
    \textbf{43.7}$^{\dagger\ddagger}$  &
    67.3 & (66.8 / 67.4)\\
    \midrule
     
    \multicolumn{7}{l}{\textbf{Improvements}}\\
    \midrule
     
    \multicolumn{2}{l}{NMT$_{\text{SRC+IMG}}$ vs. \textbf{NMT}} &
    \greenbf{$\uparrow$ 2.8} &
    \greenbf{$\uparrow$ 2.7} &
    \greenbf{$\downarrow$ 3.0} &
    \greenbf{$\uparrow2.1$} &
    \red{$\downarrow0.9$} / \green{$\uparrow2.4$} \\
    
    \multicolumn{2}{l}{NMT$_{\text{SRC+IMG}}$ vs. \textbf{PBSMT}} &
    \greenbf{$\uparrow$ 3.6} &
    \greenbf{$\uparrow0.7$} &
    \greenbf{$\downarrow$ 1.4} &
    \redbf{$\downarrow0.1$} &
    \green{$\uparrow0.3$} / \red{$\downarrow0.1$} \\
    
    \multicolumn{2}{l}{NMT$_{\text{SRC+IMG}}$ vs. Huang} &
    \greenbf{$\uparrow 1.4$} &
    \greenbf{$\uparrow 2.8$} &
    --- & --- & --- \\
    
    \multicolumn{2}{l}{NMT$_{\text{SRC+IMG}}$ vs. Huang (+RCNN)} &
    \greenbf{$\uparrow$ $0.0$} &
    \greenbf{$\uparrow$ $0.9$} &
    --- & --- & --- \\
    \midrule
    
    \multicolumn{7}{c}{\textbf{Pre-training data set: back-translated M30k$_\text{C}$} (in-domain)}\\
    \midrule
     
    
    PBSMT (LM) &
    M30k$_\text{T}$ &
    34.0 {\small \whitebf{$\uparrow0.0$}} &
    \textbf{\underline{55.0}}$^\dagger$ {\small \whitebf{$\uparrow0.0$}} &
    44.7 {\small \whitebf{$\uparrow0.0$}} &
    \textbf{\underline{68.0}} &
    (66.8 / 68.1) \\
    
    NMT &
    M30k$_\text{T}$ &
    \underline{35.5}$^\ddagger$ {\small \whitebf{$\uparrow0.0$}} &
    53.4 {\small \whitebf{$\uparrow0.0$}} &
    \underline{43.3}$^\ddagger$ {\small \whitebf{$\uparrow0.0$}} &
    65.2 &
    (67.7 / 65.0) \\
    
    NMT$_{\text{SRC+IMG}}$ &
    M30k$_\text{T}$ &
    \textbf{37.1}$^\dagger$$^\ddagger$ &
    54.5$^\dagger$\whitebf{$^\ddagger$} &
    \textbf{42.8}$^\dagger$$^\ddagger$ &
    66.6 &
    (67.2 / 66.5) \\
    \midrule
    
    \multicolumn{2}{l}{\textbf{NMT$_{\text{SRC+IMG}}$ vs. best PBSMT}} &
    \greenbf{$\uparrow$ 3.1} &
    \red{$\downarrow$ 0.5} &
    \greenbf{$\downarrow$ 1.9} &
    \redbf{$\downarrow1.4$} & 
    \greenbf{$\uparrow0.4$} / \red{$\downarrow1.6$} \\
    
    \multicolumn{2}{l}{\textbf{NMT$_{\text{SRC+IMG}}$ vs. NMT}} &
    \greenbf{$\uparrow$ 1.6} &
    \greenbf{$\uparrow$ 1.1} &
    \greenbf{$\downarrow$ 0.5} &
    \greenbf{$\uparrow1.4$} &
    \red{$\downarrow0.5$} / \green{$\uparrow1.5$} \\
    
    \midrule
    \multicolumn{7}{c}{\textbf{Pre-training data set: WMT'15 English-German corpora} (general domain)}\\
    \midrule
    
    PBSMT (concat) &
    M30k$_\text{T}$ &
    32.6 &
    53.9 &
    46.1 &
    67.3 &
    (66.3 / 67.4) \\
    
    PBSMT (LM) &
    M30k$_\text{T}$ &
    32.5 &
    54.1 &
    46.0 &
    67.3 &
    (66.0 / 67.4) \\
    
    NMT &
    M30k$_\text{T}$ &
    \underline{37.8}\whitebf{$^\dagger$} {\small\whitebf{$\uparrow0.0$}} &
    \underline{56.7}\whitebf{$^\dagger$} {\small\whitebf{$\uparrow0.0$}} &
    \underline{41.0}\whitebf{$^\dagger$} {\small\whitebf{$\uparrow0.0$}} &
    \underline{69.2} &
    (69.7 / 69.1) \\
    
    NMT$_{\text{SRC+IMG}}$ &
    M30k$_\text{T}$ &
    \textbf{39.0}$^\dagger$$^\ddagger$ &
    \textbf{56.8}\whitebf{$^\dagger$}$^\ddagger$ &
    \textbf{40.6}\whitebf{$^\dagger$}$^\ddagger$ &
    \textbf{69.6} &
    (69.6 / 69.6) \\
    \midrule
    
    \multicolumn{2}{l}{\textbf{NMT$_{\text{SRC+IMG}}$ vs. best PBSMT}} &
    \greenbf{$\uparrow$ 6.4} &
    \greenbf{$\uparrow$ 2.7} &
    \greenbf{$\downarrow$ 5.4} &
    \greenbf{$\uparrow2.3$} &
    \greenbf{$\uparrow3.3$} / \green{$\uparrow2.2$} \\
    
    \multicolumn{2}{l}{\textbf{NMT$_{\text{SRC+IMG}}$ vs. NMT}} &
    \greenbf{$\uparrow$ 1.2} &
    \green{$\uparrow$ 0.1} &
    \green{$\downarrow$ 0.4} &
    \green{$\uparrow$ 0.4} &
    \red{$\downarrow0.1$} / \green{$\uparrow0.5$} \\
    \bottomrule
  \end{tabular}
  }
  
  \caption{BLEU$4$, METEOR, chrF3, character-level precision and recall (higher is better) and TER scores (lower is better) on the translated Multi30k (M30k$_\text{T}$) test set.
  Best text-only baselines results are underlined and best overall results appear in bold.
  We show \newcite{Huangetal2016}'s improvements over the best text-only baseline in parentheses.
  Results are significantly better than the NMT baseline ($^\dagger$) and the SMT baseline ($^\ddagger$) with $p<0.01$ (no pre-training) or $p<0.05$ (when pre-training either on the back-translated M30k$_\text{C}$ or WMT'15 corpora). Best viewed in colour.
  }
  \label{tbl:results}
\end{table*}

Our encoder is a bidirectional RNN with GRU, one 1024D single-layer forward and one 1024D single-layer backward RNN.
Source and target word embeddings are 620D each and trained jointly with the model.
Word embeddings and other non-recurrent matrices are initialised by sampling from a Gaussian $\mathcal{N} (0, 0.01^2)$, recurrent matrices are random orthogonal and bias vectors are all initialised to zero.

Visual features are obtained by feeding images to the pre-trained
ResNet-50 and using the activations of the \texttt{res4f} layer~\cite{He2015}.
We apply dropout with a probability of $0.5$ in the encoder bidirectional RNN, the image features, the decoder RNN and before emitting a target word.
We follow~\newcite{Gal2016} and apply dropout to the encoder bidirectional and the decoder RNN using one same mask in all time steps.

All models are trained using
stochastic gradient descent with ADADELTA~\cite{Zeiler2012} with minibatches of size 80 (text-only NMT) or 40 (MNMT), where each training instance consists of one English sentence, one German sentence and one image (MNMT).
We apply early stopping for model selection based on BLEU4, so that if a model does not improve on BLEU4 in the validation set for more than 20 epochs, training is halted.

The translation quality of our models is evaluated quantitatively in terms of BLEU$4$~\cite{Papinenietal2002}, METEOR~\cite{DenkowskiLavie2014}, TER~\cite{Snoveretal2006}, and chrF3~\cite{Popovic2015}.\footnote{We specifically compute character 6-gram F3, and additionally character precision and recall for comparison.}
We report statistical significance with approximate randomisation for the first three metrics using the \mbox{MultEval} tool~\cite{Clarketal2011}.

\subsection{Baselines}\label{sec:baselines}

We train a text-only phrase-based SMT (PBSMT) system and a text-only NMT model for comparison.
Our PBSMT baseline is built with Moses and uses a 5--gram LM with modified Kneser-Ney smoothing~\cite{KneserNey1995}.
It is trained on the English--German descriptions of the M30k$_\text{T}$, whereas its LM is trained on the German descriptions only.
We use minimum error rate training to tune the model~\cite{Och2003} with BLEU.
The text-only NMT baseline is the one described in~\cref{sec:model} and is trained on the M30k$_\text{T}$'s English--German descriptions.

We also compare our model against two multi-modal attention-based NMT models.
The first model is \newcite{Huangetal2016}'s best model trained on the same data, and the second is their best model using additional object detections, respectively models \texttt{m1} (image at head) and \texttt{m3} in the authors' paper.

\subsection{Results}
\label{sec:results}

In Table~\ref{tbl:results}, we show results for our text-only baselines NMT and PBSMT, the multi-modal models of \newcite{Huangetal2016} and our MNMT models trained on the M30k$_\text{T}$, and pre-trained on the in-domain back-translated M30k$_\text{C}$ and the general-domain text-only English-German MT corpora from WMT 2015.

\paragraph{Training on M30k$_\text{T}$}

One main finding is that our model consistently outperforms the comparable model of \newcite{Huangetal2016}, with improvements of $+1.4$ BLEU and $+2.7$ METEOR.
In fact, even when their model has access to more data our model still improves by $+0.9$ METEOR, while maintaining the same BLEU4 scores.

Moreover, we can also conclude from Table~\ref{tbl:results} that PBSMT performs better at recall-oriented metrics, i.e. METEOR and chrF3, whereas NMT is better at precision-oriented ones, i.e. BLEU4.
This is somehow expected, since the attention mechanism in NMT~\cite{BahdanauChoBengio2015} does not explicitly take attention weights from previous time steps into account, an thus lacks the notion of source coverage as in SMT~\cite{Koehnetal2003,Tuetal2016}.
We note that these ideas are complementary and incorporating coverage into model NMT$_{\text{SRC+IMG}}$ could lead to more improvements, especially in recall-oriented metrics.
Nonetheless, our doubly-attentive model shows consistent gains in both precision- and recall-oriented metrics in comparison to the text-only NMT baseline, i.e. it is significantly better according to BLEU4, METEOR and TER ($p<0.01$), and it also improves chrF3 by $+2.1$.
In comparison to the PBSMT baseline, our proposed model still significantly improves according to both BLEU4 and TER ($p<0.01$), also increasing METEOR by $+0.7$ but with an associated $p$-value of $p=0.071$, therefore not significant for $p<0.05$.
Although chrF3 is the only metric in which the PBSMT model scores best, the difference between our model and the latter is only $0.1$, meaning that they are practically equivalent.
We note that model NMT$_{\text{SRC+IMG}}$ consistently increases character recall in comparison to the text-only NMT baseline.
Although it can happen at the expense of character precision, gains in recall are always much higher than any eventual loss in precision, leading to consistent improvements in chrF3.

\begin{figure*}[ht!]
 \centering
 \begin{subfigure}{0.45\linewidth}
   \hfill
   \includegraphics[width=\textwidth]{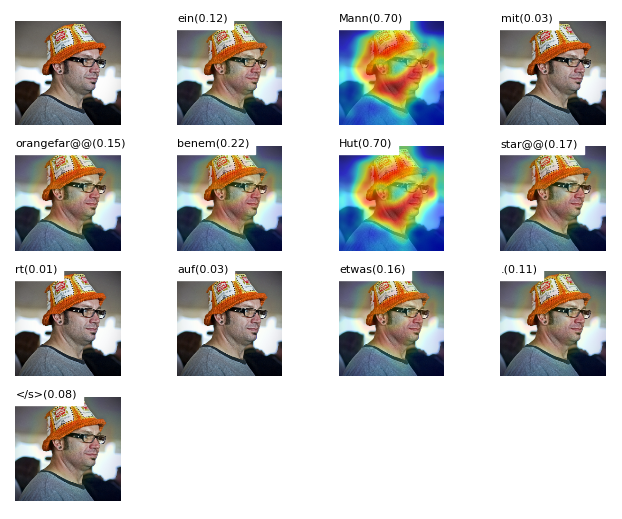}
   \caption{Image--target word alignments.}
 \end{subfigure}
 \begin{subfigure}{0.54\linewidth}
   \centering
   \includegraphics[width=\textwidth]{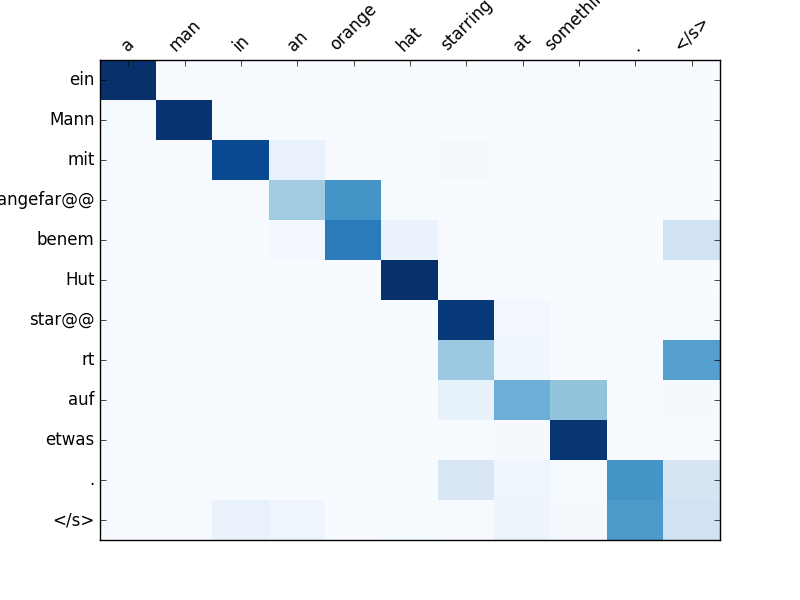}
   \caption{Source--target word alignments.}
 \end{subfigure}
 \caption{Visualisation of image-- and source--target word alignments for the M30k$_\text{T}$ test set.}
 \label{fig:attention}
\end{figure*}

\paragraph{Pre-training}

We now discuss results for models pre-trained using different data sets.
We first pre-trained the two text-only baselines PBSMT and NMT, and our MNMT model on the back-translated M30k$_\text{C}$, a medium-sized in-domain image description data set ($145$k training instances).
We also pre-trained the same models on the English--German parallel sentences of much larger MT data sets, i.e. the concatenation of the Europarl~\cite{Koehn2005}, Common Crawl and News Commentary corpora, used in WMT 2015 ($\sim$$4.3$M parallel sentences).
Model PBSMT (concat.) used the concatenation of the pre-training and training data for training, and model PBSMT (LM) used the general-domain German sentences as additional data to train the LM.
From Table \ref{tbl:results}, it is clear that model NMT$_{\text{SRC+IMG}}$ can learn from both in-domain, multi-modal pre-training data sets as well as text-only, general domain ones.

\paragraph{Pre-training on M30k$_\text{C}$}

When pre-training on the back-translated M30k$_\text{C}$, the recall-oriented chrF3 shows a difference of $1.4$ points between PBSMT and our model, mostly due to character recall; nonetheless, our model still improved by the same margin on the text-only NMT baseline.
Our model still outperforms the PBSMT baseline according to BLEU4 and TER, and the text-only NMT baseline according to all metrics ($p<.05$).

\paragraph{Pre-training on WMT 2015 corpora}

We also pre-trained our models on the WMT 2015 corpora, which took 10 days, i.e. $\sim$6--7 epochs.
Results show that model NMT$_{\text{SRC+IMG}}$ improves significantly over the NMT baseline according to BLEU4, and is consistently better than the PBSMT baseline according to all four metrics.\footnote{In order for PBSMT models to remain competitive, we believe more advanced data selection techniques are needed, which are out of the scope of this work.}
This is a strong indication that model NMT$_{\text{SRC+IMG}}$ can exploit the additional pre-training data efficiently, both general- and in-domain.
While the PBSMT model is still competitive when using additional in-domain data---according to METEOR and chrF3--- the same cannot be said when using general-domain pre-training corpora.
From our experiments, NMT models in general, and especially model NMT$_{\text{SRC+IMG}}$, thrive when training and test domains are mixed, which is a very common real-world scenario.

\paragraph{Textual and visual attention}

In Figure~\ref{fig:attention}, we visualise the visual and textual attention weights for an entry of the M30k$_\text{T}$ test set.
In the visual attention, the $\beta$ gate (written in parentheses after each word) caused the image features to be used mostly to generate the words \emph{Mann} (man) and \emph{Hut} (hat), two highly visual terms in the sentence.
We observe that in general visually grounded terms, e.g. Mann and Hut, usually have a high associated $\beta$ value, whereas other less visual terms like \emph{mit} (with) or \emph{auf} (at) do not.
That causes the model to use the image features when it is describing a visual concept in the sentence, which is an interesting feature of our model.
Interestingly, our model is very selective when choosing to use image features: it only assigned $\beta>0.5$ for $20\%$ of the outputted target words, and $\beta>0.8$ to only $8\%$.
A manual inspection of translations shows that these words are mostly concrete nouns with a strong visual appeal.

Lastly, using two independent attention mechanisms is a good compromise between model compactness and flexibility.
While the attention-based NMT model baseline has $\sim$200M parameters,
model NMT$_{\text{SRC+IMG}}$ has $\sim$213M,
thus using just $\sim$6.6\% more parameters than the latter.

\section{Related work}
\label{sec:related}

Multi-modal MT was just recently addressed by the MT community by means of a shared task~\cite{Speciaetal2016}.
However, there has been a considerable amount of work on natural language generation from non-textual inputs.
\newcite{Maoetal2014} introduced a multi-modal RNN that integrates text and visual features and applied it to the tasks of image description generation and image--sentence ranking.
In their work, the authors incorporate global image features in a separate multi-modal layer that merges the RNN textual representations and the global image features.
\newcite{Vinyalsetal2015} proposed an influential neural IDG model based on the sequence-to-sequence framework, which is trained end-to-end.
\newcite{Elliottetal2015} put forward a model to generate multilingual descriptions of images by learning and transferring features between two independent, non-attentive neural image description models.\footnote{Although their model has not been devised with translation as its primary goal, theirs is one of the baselines of the first shared task in multi-modal MT in WMT 2016~\cite{Speciaetal2016}.}
\newcite{Venugopalanetal2015} introduced a model trained end-to-end to generate textual descriptions of open-domain videos from the video frames based on the sequence-to-sequence framework.
Finally, \newcite{Xuetal2015} introduced the first attention-based IDG model where an attentive decoder learns to attend to different parts of an image as it generates its description in natural language.

In the context of NMT, \newcite{Dongetal2015} proposed a multi-task learning approach where a model is trained to translate from one source language into multiple target languages.
They used attention-based decoders where each language has one decoder RNN with a separate attention mechanism.
Each translation task has a shared source-language encoder in common with all the other translation tasks.
\newcite{Firatetal2016} proposed a multi-way model trained to translate between many different source and target languages.
Instead of one attention mechanism per language pair as in~\newcite{Dongetal2015}, which would lead to a quadratic number of attention mechanisms in relation to language pairs, they use a shared attention mechanism where each target language has one attention shared by all source languages.
\newcite{Luongetal2016} proposed a multi-task approach where they train a model using two tasks and a shared decoder: the main task is to translate from German into English and the secondary task is to generate English image descriptions.
They show improvements in the main translation task when also training for the secondary image description task.
Although not an NMT model, \newcite{Hitschleretal2016} recently used image features to re-rank translations of image descriptions generated by an SMT model and reported significant improvements.

Although no purely neural multi-modal model to date significantly improves on both text-only NMT and SMT models~\cite{Speciaetal2016}, different research groups have proposed to include global and spatial visual features in re-ranking $n$-best lists generated by an SMT system or directly in an NMT framework with some success~\cite{Caglayanetal2016,CalixtoElliottFrank2016,Huangetal2016,Libovickyetal2016,Shahetal2016}.
To the best of our knowledge, the best published results of a purely MNMT model are those of \newcite{Huangetal2016}, who proposed to use global visual features extracted with the VGG19 network~\cite{SimonyanZisserman2014} for an entire image, and also for regions of the image obtained using the RCNN of \newcite{Girshicketal2014}.
Their best model improves over a strong text-only NMT baseline and is comparable to results obtained with an SMT model trained on the same data.
For that reason, their models are used as baselines in our experiments.

Our work differs from previous work in that, first, we propose attention-based MNMT models.
This is an important difference since the use of attention in NMT has become standard and is the current state-of-the-art~\cite{Jeanetal2015,Luongetal2015,Firatetal2016,Sennrichetal2016}.
Second, we propose a \emph{doubly-attentive model} where we effectively fuse two mono-modal attention mechanisms into one multi-modal decoder, training the entire model jointly and end-to-end.
In addition, we are interested in how to merge textual and visual representations into multi-modal representations when generating words in the target language, which differs substantially from text-only translation tasks even when these translate from many source languages into many target languages \cite{Dongetal2015,Firatetal2016}.
To the best of our knowledge, we are the first to integrate multi-modal inputs in NMT via independent attention mechanisms.

\section{Conclusions and Future Work}\label{sec:conclusions}

We have introduced a novel attention-based, multi-modal NMT model to incorporate spatial visual information into NMT.
We have reported new state-of-the-art results on the M30k$_\text{T}$ test set, improving on previous multi-modal attention-based models.
We also pre-trained our model on one in-domain multi-modal data set and many general-domain text-only MT corpora, finding that it learns efficiently and is able to exploit the additional data regardless of the domain.
Our model also compares favourably to both NMT and PBSMT baselines evaluated on the same training data.

In the future, we will incorporate coverage into our model and study how to apply it to other Natural Language Processing tasks.

\bibliographystyle{acl_natbib}
\bibliography{phd}

\end{document}